%% file: main.tex
\documentclass[runningheads]{llncs}
\usepackage{graphicx}

\usepackage{tikz}
\usepackage{comment}
\usepackage{amsmath,amssymb} %
\usepackage{color}

\usepackage[accsupp]{axessibility}  %

\usepackage[width=122mm,left=12mm,paperwidth=146mm,height=193mm,top=12mm,paperheight=217mm]{geometry}

\usepackage{graphicx}
\usepackage{amsmath}
\usepackage{amssymb}
\usepackage{booktabs}
\usepackage{tabularx}
\usepackage[export]{adjustbox}
\usepackage{float}
\usepackage{arydshln}
\usepackage{tablefootnote}
\usepackage{subfig}
\usepackage{wrapfig}

\usepackage[pagebackref,breaklinks,colorlinks]{hyperref}

\usepackage[capitalize]{cleveref}
\crefname{section}{Sec.}{Secs.}
\Crefname{section}{Section}{Sections}
\Crefname{table}{Table}{Tables}
\crefname{table}{Tab.}{Tabs.}

\newcommand{\Table}[1]{Table~\ref{tab:#1}}

\newif\ifdrafting
\draftingfalse %
\ifdrafting
    \newcommand{\ds}[1]{{\color{red}[DS: #1]}}
    \newcommand{\fr}[1]{{\color{blue}[FR: #1]}}
    \newcommand{\newedit}[1]{ {\color{cyan} {#1}}}
\else
	\newcommand{\ds} [1] {}
	\newcommand{\fr}[1]{}
	\newcommand{\newedit}[1]{{#1}}
\fi

\begin{document}
\captionsetup[figure]{labelfont={bf},labelformat={default},labelsep=period,name={Fig.}}
\captionsetup[table]{labelfont={bf},labelformat={default},labelsep=period,name={Table}}

\pagestyle{headings}
\mainmatter
\def\ECCVSubNumber{4614}  %

\title{FILM: Frame Interpolation for Large Motion} %

\titlerunning{FILM: Frame Interpolation for Large Motion}
\author{Fitsum Reda\inst{1} \and
Janne Kontkanen\inst{1} \and
Eric Tabellion\inst{1} \and
Deqing Sun\inst{1} \and \\
Caroline Pantofaru\inst{1} \and
Brian Curless\inst{1,2}}
\authorrunning{F. Reda et al.}
\institute{\hspace{-1ex}Google Research \and \hspace{-1ex}University of Washington}
\maketitle
\begin{abstract}
We present a frame interpolation algorithm that synthesizes an engaging slow-motion video from near-duplicate photos which often exhibit large scene motion. Near-duplicates interpolation is an interesting new application, but large motion poses challenges to existing methods. To address this issue, we adapt a feature extractor that shares weights across the scales, and present a ``scale-agnostic'' motion estimator. It relies on the intuition that large motion at finer scales should be similar to small motion at coarser scales, which boosts the number of available pixels for large motion supervision. To inpaint wide disocclusions caused by large motion and synthesize crisp frames, we propose to optimize our network with the Gram matrix loss that measures the correlation difference between features. To simplify the training process, we further propose a unified single-network approach that removes the reliance on additional optical-flow or depth network and is trainable from frame triplets alone. Our approach outperforms state-of-the-art methods on the Xiph large motion benchmark while performing favorably on Vimeo-90K, Middlebury and UCF101. Source codes and pre-trained models are available at \url{https://film-net.github.io}.

\end{abstract}

\input{sec/1_introduction}
\input{sec/2_related}
\input{sec/3_method}
\input{sec/4_implementation}
\input{sec/5_results}
\input{sec/6_conclusions}

\clearpage

\noindent\textbf{Acknowledgements.} We thank Orly Liba and Charles Herrmann for feedback on the text, Jamie Aspinall for the imagery in the paper, Dominik Kaeser, Yael Pritch, Michael Nechyba and David Salesin for support.

\bibliographystyle{splncs04}
\bibliography{main}

\input{supplementary}

\end{document}

%% file: sec/1_introduction.tex
\section{Introduction}
\label{sec:intro}
Frame interpolation -- synthesizing intermediate images between a pair of input frames -- is an important problem with increasing reach. It is often used for temporal up-sampling to increase refresh rate or create slow-motion videos.

Recently, a new use case has emerged. Digital photography, especially with the advent of smartphones, has made it effortless to take several pictures within a few seconds, and people naturally do so often in their quest for just the right photo that captures the moment. These ``near duplicates'' create an exciting opportunity: interpolating between them can lead to surprisingly engaging videos that reveal scene (and some camera) motion, often delivering an even more pleasing sense of the moment than any one of the original photos. 

\newedit{Unlike video, however, the temporal spacing between near duplicates can be a second or more, with commensurately large scene motion, posing a major challenge for existing interpolation methods.
Frame interpolation between consecutive video frames, which often exhibit small motion, has been studied extensively, and recent methods~\cite{softsplat-2020,dain-2019,abme-2021,huang2020rife} show impressive results for this scenario. However, little attention has been given to interpolation for large scene motion, commonly present in near duplicates. The work of~\cite{eXVFI-2021} attempted to tackle the large motion problem by training on an extreme motion dataset, but its effectiveness is limited when tested on small motion~\cite{abme-2021}.}
\begin{figure*}[t!]
    \vspace{-1.5ex}
    \centering
    \includegraphics[width=0.8\textwidth,trim=248 67.5 15.5 22,clip]{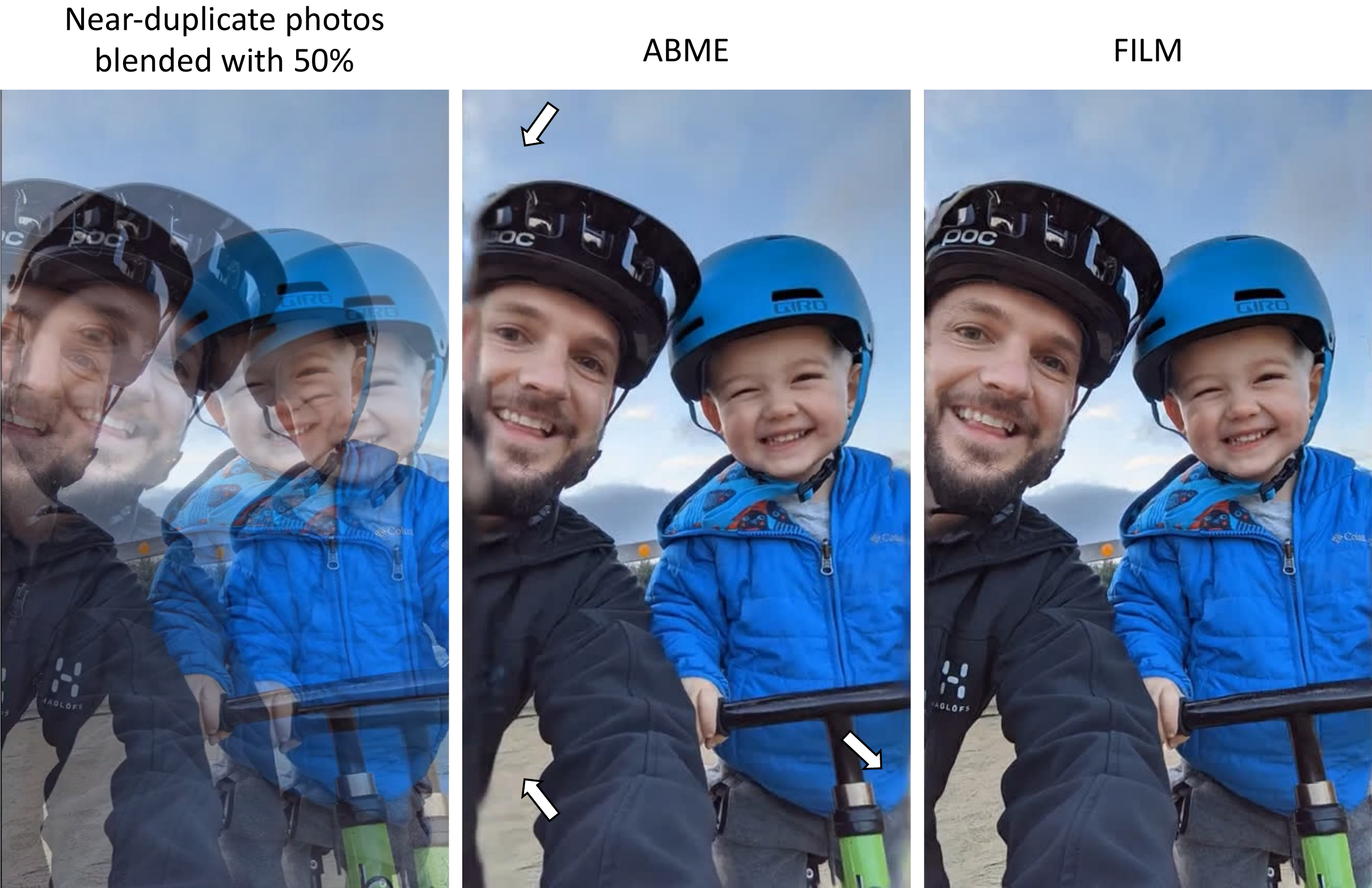}
    \caption{Near-duplicate photos interpolation with ABME~\cite{abme-2021}, showing large artifacts, and our FILM, showing improvements.}
    \label{fig:near_duplicates}
    \vspace{-4ex}
\end{figure*}

In this work, we instead propose a network that generalizes well to both small and large motion. Specifically, we adapt a multi-scale feature extractor from~\cite{fusion-2019} that shares weights across the scales and present a ``scale-agnostic'' bi-directional motion estimation module. Our approach relies on the intuition that large motion at finer scales should be similar to small motion at coarser scales, thus increasing the number of pixels (as finer scale is higher resolution) available for large motion supervision. We found this approach to be surprisingly effective in handling large motion by simply training on regular frames  (see Figure~\ref{fig:near_duplicates}).

We also observed that, while the state-of-the-art methods score well on benchmarks~\cite{ucf101,middlebury,vimeo}, the interpolated frames often appear blurry, especially in large disoccluded regions that arise from large motions. %
Here, we propose to optimize our models with the Gram matrix loss, which matches the {\em auto-correlation} of the high-level VGG features, and significantly improves the realism and sharpness of frames (see Figure~\ref{fig:disocclusions}). %

\newedit{Another drawback of recent interpolation methods~\cite{softsplat-2020,dain-2019,abme-2021,zhang2020flexible} is training complexity, because they typically rely on scarce data to pre-train additional optical flow, depth, or other prior networks.} Such data scarcity is even more critical for large motion. %
The DAIN approach~\cite{dain-2019}, for example, incorporates a depth network, and the works in~\cite{softsplat-2020,abme-2021} use additional networks to estimate per-pixel motion. 
\newedit{To simplify the training process, another contribution of this work is a unified architecture for frame interpolation, which is trainable from regular frame triplets alone.}

In summary, the main contributions of our work are:
\begin{itemize}
  \item We expand the scope of frame interpolation to a novel near-duplicate photos interpolation application, and open a new space for the community to tackle.  
  \item We adapt a multi-scale feature extractor that shares weights, and propose a scale-agnostic bi-directional motion estimator to handle both small and large motion well, using regular training frames. 
  \item We adopt a Gram matrix-based loss function to inpaint large disocclusions caused by large scene motion, leading to crisp and pleasing frames. %
  \item We propose a unified, single-stage architecture, to simplify the training process and remove the reliance on additional optical flow or depth networks. %
\end{itemize}

%% file: sec/2_related.tex
\section{Related Work}
\label{sec:related}
Various CNN-based frame interpolation methods~\cite{superslomo-2017,dain-2019,sepconv-2017,softsplat-2020,cdfi-2021,huang2020rife,zhang2020flexible,adacof-2020,bmbc-2021,abme-2021,zooming-slomo}\\~\cite{reda-cycle,cyclic-gen} have been proposed to up-scale frame rate of videos. To our knowledge, no prior work exists on near-duplicate photos interpolation. We, however, summarize frame interpolation methods related to our approach.%

\vspace{0.05in}
\noindent\textbf{Large Motion.} Handling large motion is an important yet under-explored topic in frame interpolation. The work in~\cite{eXVFI-2021} handles large motion by training on 4K sequences with extreme motion. While this is a viable approach, it does not generalize well on regular footage as discussed in~\cite{abme-2021}. Similarly, other approaches~\cite{softsplat-2020,abme-2021} perform poorly when the test motion range deviates from the training motion range. We adapt a multi-scale shared feature extractor~\cite{fusion-2019,shared-optical-flow1,shared-optical-flow2}, and present a ``scale-agnostic'' motion estimation module, which allows us to learn large and small motion with equal priority, and show favorable generalization ability in various benchmarks. %

\vspace{0.05in}
\noindent\textbf{Image Quality.} One of our key contributions is high quality frame synthesis, especially in large disoccluded regions caused by large motion. Prior work~\cite{sepconv-2017,sepconv-orig,revisiting-sepconv} improves image quality by learning a per-pixel kernel instead of an offset vector, which is then convolved with the inputs. While effective at improving quality, they cannot handle large motion well. Other approaches optimize models with perceptual losses~\cite{sepconv-2017,softsplat-2020}. Some consider an adversarial loss~\cite{deep-gan-loss}, albeit with a complex training process. Our work proposes to adopt the Gram matrix loss~\cite{style-transfer}, which builds up on the perceptual loss and yields high quality and pleasing frames. 

\vspace{0.05in}
\noindent\textbf{Single-Stage Networks.} The first CNN-based supervised frame interpolators propose UNet-like networks, trained from inputs and target frames~\cite{ucf101,superslomo-2017}. Recent work~\cite{dain-2019} introduces a depth network to handle occlusions, \cite{softsplat-2020,abme-2021} incorporate motion estimation modules, and \cite{sun2018pwc,bmbc-2021,zhang2020flexible} rely on pre-trained HED~\cite{xie15hed} features. While impressive results are achieved, multiple networks can make training processes complex. It may also need scarce data to pre-train the prior networks. Pre-training datasets, e.g. optical flows, are even more scarce for large motion. Our work introduces a single unified network, trainable from regular frame triples alone, without additional priors, and achieves state-of-the-art results.

%% file: sec/3_method.tex
\section{Method}
\label{sec:method}
Given two input images $(\textbf{I}_{0}, \textbf{I}_{1})$, with large in-between motion, we synthesize a mid-image $\hat{\textbf{I}}_{t}$, with time  $t\in(0,1)$, as:
\begin{equation}\label{eq:film_model}
\hat{\textbf{I}}_{t} = \mathcal{M}{\big(\textbf{I}_{0}, \textbf{I}_{1}\big)} ,
\end{equation}
where $\mathcal{M}$ is our FILM network trained with a ground-truth $\textbf{I}_{t}$. During training, we supervise at $t=0.5$ and we predict more in-between images by recursively invoking FILM.

A common approach to handle large motion is to employ feature pyramids, which increases receptive fields. However, a standard pyramid learning has two difficulties: 1) small fast-moving objects disappear at coarse levels, and 2) the number of pixels is drastically smaller at coarse levels ($i$), $(\frac{H}{2^{i}}\mathord\times\mathord\frac{W}{2^{i}})$, which means there are fewer pixels to provide large motion supervision. To overcome these challenges, we propose to share the convolution weights across the scales. Based on the intuition that large motion at finer scales should be the same as small motion at coarser scales, sharing weights allows us to boost the number of pixels available for large motion supervision.

\begin{figure*}[t!]
    \centering
    \includegraphics[width=0.9\textwidth]{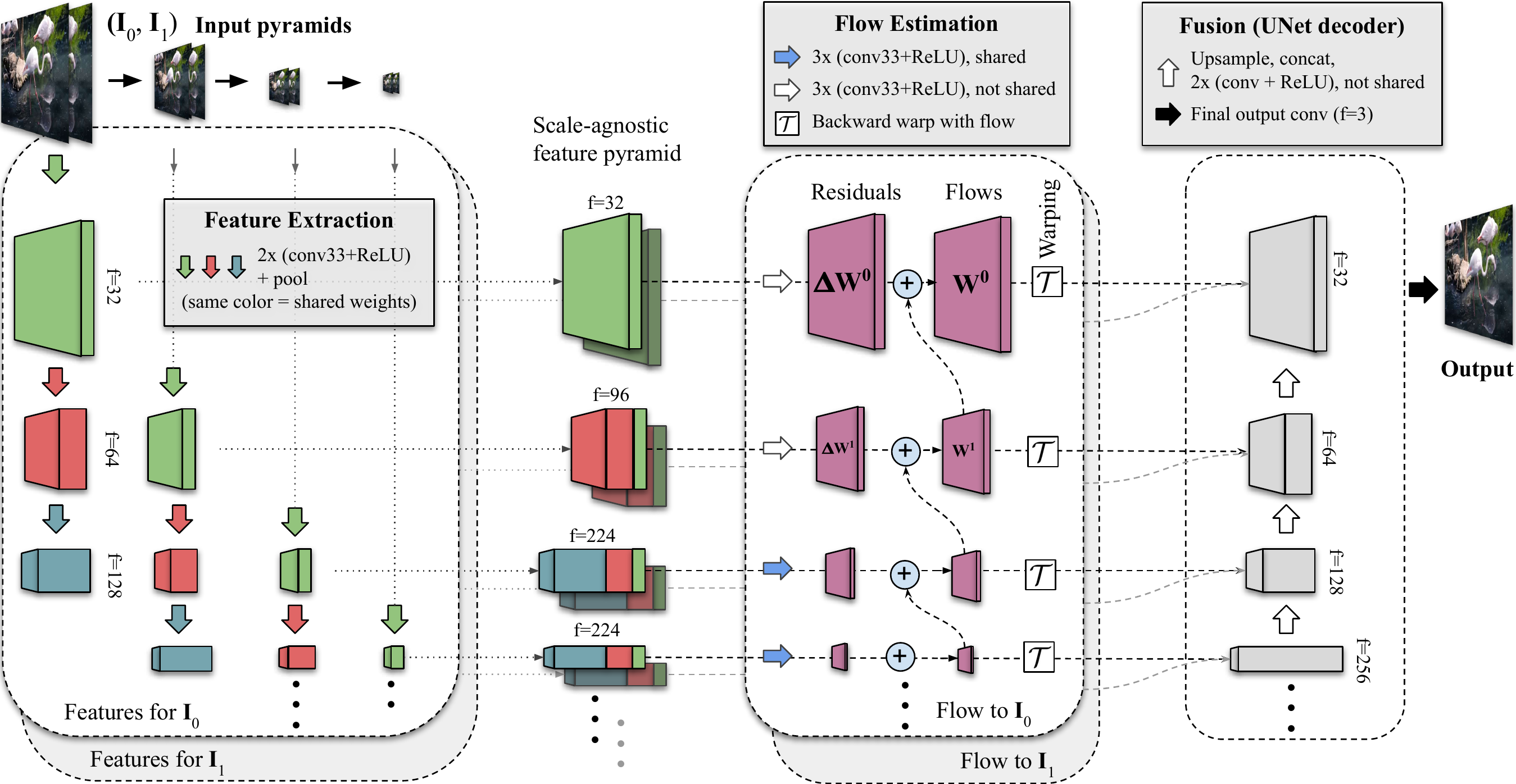}
    \caption{FILM architecture. Our flow estimation module computes ``scale agnostic'' bi-directional flows based on the feature pyramids, extracted by shared weights.}
    \label{fig:architecture}
    \vspace{-4ex}
\end{figure*}

FILM has three main stages: Shared feature extraction, scale-agnostic motion estimation, and a fusion stage that outputs the resulting color image. Figure~\ref{fig:architecture} shows an overview of FILM.

\vspace{0.05in}
\noindent\textbf{Feature Extraction.} %
We adapt a feature extractor from~\cite{fusion-2019}, that allows weight sharing across the scales, to create a ``scale-agnostic'' feature pyramid. It is constructed in three steps as follows.
 
First, we create image pyramids $\{\textbf{I}_0^l\}$ and $\{\textbf{I}_1^l\}$ for the two input images, where $l\in[1,7]$ is the pyramid level. 

Second, starting at the image at each $l$-th pyramid level, we build feature pyramids (the columns in Figure~\ref{fig:architecture}) using a shared UNet encoder. Specifically, we extract multi-scale features $\{\textbf{f}_0^{l,d}\}$ and $\{\textbf{f}_1^{l,d}\}$, with $d\in[1,4]$ being the depth at that $l$-th level (Figure~\ref{fig:architecture} only uses $d\in[1,3]$ for illustration). Mathematically,
\begin{equation}\label{eq:subnet}
\textbf{f}_0^{l,d} = \mathcal{H}^{d}{\big(\textbf{I}_{0}^{l}\big)}, 
\end{equation}
where $\mathcal{H}^{d}$ is a stack of convolutions, shown in Figure~\ref{fig:architecture} with the green arrow for $d\mathord=1$, red for $d\mathord=2$, and dark-blue for $d\mathord=3$. Note that, the same $\theta^{(\mathcal{H}^{d})}$ convolution weights are shared for the same $d$-th depth at each pyramid level, to create {\em compatible multi\-scale features}. Each $\mathcal{H}^{d}$ is followed by an average pooling with a size and stride of 2.

As a third and final step of our feature extractor, we construct our scale-agnostic feature pyramids, $\{\textbf{F}_0^{l}\}$ and $\{\textbf{F}_1^{l}\}$, by concatenating the feature maps with different depths, but the same spatial dimensions, as:
\begin{equation}\label{eq:scale_agnostic}
\textbf{F}_0^{l} = \Big(\textbf{f}_0^{l-2,d=3}, \textbf{f}_0^{l-1,d=2}, \textbf{f}_0^{l,d=1}\Big) , 
\end{equation}
and the scale-agnostic feature, $\textbf{F}_1^{l}$ of $\textbf{I}_{1}$, at the $l$-th pyramid level, can be given in a similar way by Equation~\ref{eq:scale_agnostic}. As shown in Figure~\ref{fig:architecture}, the finest level feature (green) can only aggregate one feature map, the second finest level two (green+red), and the rest can aggregate three shared feature maps.

\vspace{0.05in}
\noindent\textbf{Flow Estimation.} Once we extract the feature pyramids, $\{\textbf{F}_0^{l}\}$ and $\{\textbf{F}_1^{l}\}$, we use them to calculate a bi-directional motion at each pyramid level. Similar to~\cite{sun2018pwc}, we start the motion estimation from the coarsest level (in our case $l=7$). However, in contrast to other methods, we directly predict task oriented~\cite{vimeo,ucf101} flows, $\textbf{W}_{t\rightarrow 0}$ and $\textbf{W}_{t\rightarrow 1}$, from the mid-frame to the inputs.

We compute the task oriented flow at each level $\textbf{W}_{t\rightarrow1}^{l}$ as the sum of predicted residual and the upsampled flow from the coarser level $l+1$, based on the intuition that large motion at finer scales should be the same as small motion at coarser scales, as:
\begin{equation}\label{eq:flow_to_interm}
\textbf{W}_{t\rightarrow1}^{l} = \big(\textbf{W}_{t\rightarrow1}^{l+1}\big)_{\times2} + \mathcal{G}^{l}\Big(\textbf{F}_{0}^{l}, \hat{\textbf{F}}_{t \leftarrow 1}^{l}\Big) , 
\end{equation}
where $(\bullet)_{\times2}$ is a bilinear up-sampling, $\mathcal{G}^{l}$ is a stack of convolutions that estimates the residual, and $\hat{\textbf{F}}_{t\leftarrow1}^{l}$ is the backward warped scale-agnostic feature map at $t\mathord=1$, obtained by bilinearly warping $\textbf{F}_{1}^{l}$ with the upsampled flow estimate, as,   
\begin{equation}\label{eq:residual_flow_forward}
\hat{\textbf{F}}_{t\leftarrow1}^{l} = \mathcal{T}\Big(\textbf{F}_1^{l}, \big(\textbf{W}_{t\rightarrow1}^{l+1}\big)_{\times2}\Big) , 
\end{equation} 
with $\mathcal{T}$ being a bilinear resample (warp) operation. Figure~\ref{fig:architecture} depicts $\mathcal{G}^{l}$ by the blue or white arrows, depending on the pyramid level. Note that, the same residual convolution weights $\theta^{(\mathcal{G}^{l})}$~are shared by levels $l\in[3,7]$.%

Finally, we create the feature pyramid at the intermediate time $t$, $\{\textbf{F}_{t\leftarrow1}^{l}\}$ and $\{\textbf{F}_{t\leftarrow0}^{l}\}$, by backward warping the feature pyramid, at $t\mathord=1$ and $t\mathord=0$, with the flows given by Equation~\ref{eq:flow_to_interm}, as:
\begin{equation}\label{eq:feature_inerm}
\textbf{F}_{t\leftarrow1}^{l} = \mathcal{T}\Big(\big(\textbf{F}_1^{l},\textbf{I}_{1}^{l}\big), \textbf{W}_{t\rightarrow 1}^{l}\Big) , 
\end{equation} 
$\textbf{F}_{t\leftarrow0}^{l}$ can be given in a similar way as Equation~\ref{eq:feature_inerm}.

\vspace{0.05in}
\noindent\textbf{Fusion.} The final stage of FILM concatenates, at each $l$-th pyramid, the scale-agnostic feature maps at $t$ and the bi-directional motions to $t$, which are then fed to a UNet-like~\cite{unet} decoder to synthesize the final mid-frame $\hat{\textbf{I}}_{t}$. Mathematically, the fused input at each $l$-th decoder level is given by, 
\begin{equation}\label{eq:feature_fusion}
\Big(\textbf{F}_{t\leftarrow1}^{l}, \textbf{F}_{t\leftarrow0}^{l}, \textbf{W}_{t\rightarrow 0}^{l}, \textbf{W}_{t\rightarrow 1}^{l}\Big).
\end{equation}
Figure~\ref{fig:architecture} illustrates the decoder's convolutions and resulting activations with a white arrow and gray boxes, respectively.

\subsection{Loss Functions}
We use only image synthesis losses to supervise the final output of our FILM network; we do not use auxiliary losses tapped into any intermediate stages. Our image synthesis loss is a combination of three terms.

First, we use the L1 reconstruction loss that minimizes the pixel-wise RGB difference between the interpolated frame $\hat{\textbf{I}}_{t}$ and the ground-truth frame $\textbf{I}_{t}$, given by:
\begin{equation}\label{eq:color}
\mathcal{L}_{1} = {\lVert\hat{\textbf{I}}_{t}-\textbf{I}_{t}\rVert}_{1}. 
\end{equation}
The $\mathcal{L}_{1}$ loss captures the motion between the inputs $(\textbf{I}_{0}, \textbf{I}_{1})$ and yields interpolation results that score well on benchmarks, as is discussed in Section~\ref{subsec:qualitative_comparisons}. However, the interpolated frames are often blurry. %

Second, to enhance image details, we add a perceptual loss, using the L1 norm of the VGG-19 features~\cite{vgg-16}. The perceptual loss, also called VGG-loss, $\mathcal{L}_{\rm VGG}$, is given by, 
\begin{equation} \label{eq:perceptual}
{\mathcal{L}_{\rm VGG}} = \frac{1}{L}\sum_{l=1}^{L}{\alpha_l {\left\lVert{\Psi}_{l}(\hat{\textbf{I}}_{t})-{\Psi}_{l}(\textbf{I}_{t})\right\rVert}_{1} }, 
\end{equation}
where ${\Psi}_{l}(\textbf{I}_{i}) \in \mathbb{R}^{\mathrm{H}\times\mathrm{W}\times\mathrm{C}}$ is the features from the ${l}$-th selected layer of a pre-trained Imagenet VGG-19 network for $\textbf{I}_{i} \in \mathbb{R}^{\mathrm{H}\times\mathrm{W}\times\mathrm{3}}$, $\mathit{L}$ is the number of the finer layers considered, and $\alpha_{l}$ is an importance weight of the $l$-th layer. %

Finally, we employ the Style loss~\cite{style-transfer,reda-sdc,liu-inpainting} to further expand on the benefits of $\mathcal{L}_{\rm VGG}$. The style loss $\mathcal{L}_{\rm Gram}$, also called Gram matrix loss, is the L2 norm of the auto-correlation of the VGG-19 features~\cite{vgg-16}: 

\begin{equation} \label{eq:gram-loss}
{\mathcal{L}_{\rm Gram}} = \frac{1}{L}\sum_{l=1}^{L}{ \alpha_l{\left\lVert{\mathrm{M}_{l}(\hat{\textbf{I}}_{t})} -  {\mathrm{M}_{l}(\textbf{I}_{t})}\right\rVert}_{2} } , 
\end{equation}
where the Gram matrix of the interpolated frame at the  $l$-th layer, $\mathrm{M}_{l}(\hat{\textbf{I}}_{t}) \in \mathbb{R}^{\mathrm{C}\times\mathrm{C}}$, is given by:
\begin{equation} \label{eq:gram-matrix}
    {\mathrm{M}_{l}(\hat{\textbf{I}}_{t})} = {\big(\Psi_{l}(\hat{\textbf{I}}_{t})\big)^{\intercal}{\big(\Psi}_{l}(\hat{\textbf{I}}_{t})\big)} ,
\end{equation}
and the Gram matrix of the ground-truth image, $\mathrm{M}_{l}(\textbf{I}_{t})$, can be given in a similar way as Equation~\ref{eq:gram-matrix}. 

To our knowledge, this is the first work that applies the Gram matrix loss to frame interpolation. We found this loss to be effective in synthesizing sharp images with rich details when inpainting large disocclusion regions caused by large scene motion. %

To achieve high benchmark scores as well as high quality frame synthesis, we train our models with an optimally weighted combination of the RGB, VGG and Gram matrix losses. The combined loss, which we denote $\mathcal{L}_{S}$, is defined as, 
\begin{equation}\label{eq:style}
    \mathcal{L}_{S}=w_l\mathcal{L}_1+w_{\rm VGG}\mathcal{L}_{\rm VGG}+w_{\rm Gram}\mathcal{L}_{\rm Gram} , 
\end{equation}
with the weights $(w_{l}, w_{\rm VGG}, w_{\rm Gram})$ determined empirically, as detailed in the Supplementary Materials.%

\subsection{Large Motion Datasets}
\label{sec:training_datasets}
To study FILM’s ability to handle large motion, we created a ``bracketed'' dataset containing five training sub-sets. Each containing examples with motion disparity in the following ranges, in pixels: 0-40, 0-60, 0-80, 0-100, and 0-120.

\begin{figure*}[ht!]
    \hspace{4ex}
    \centering
    \subfloat[\centering Vimeo motion magnitude.\label{fig:sub1}]{{\includegraphics[width=0.22\textwidth,trim=30 2 30 50,clip]{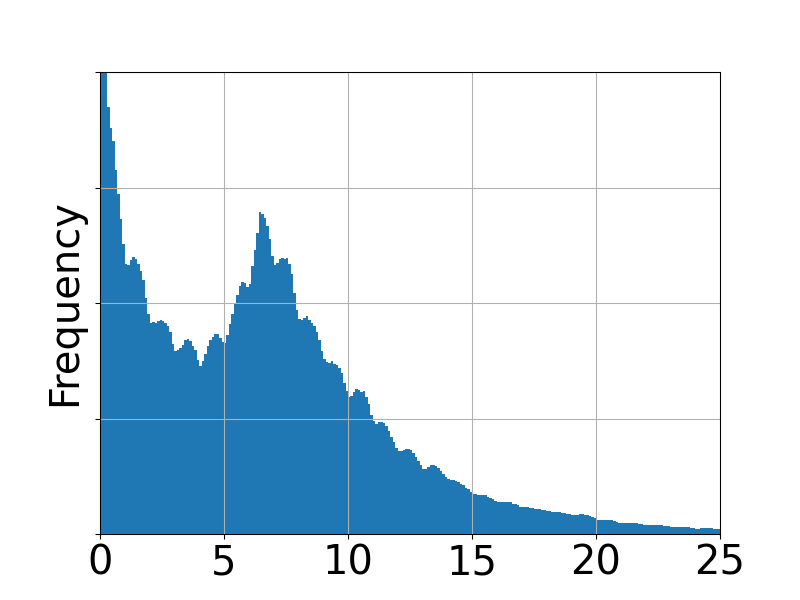} }}%
    \enspace
    \subfloat[\centering Xiph motion magnitude.\label{fig:sub2}]{{\includegraphics[width=0.22\textwidth,trim=30 2 30 50,clip]{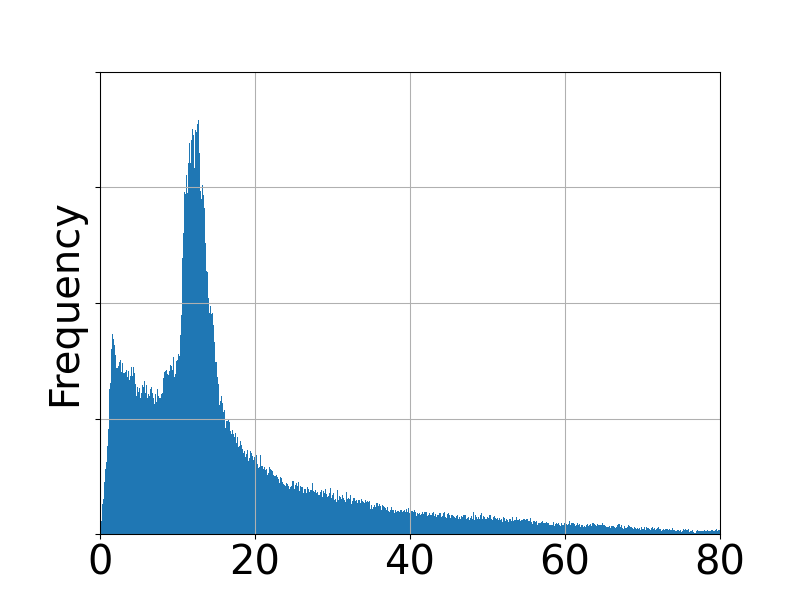} }}%
    \enspace
    \subfloat[\centering Bracketed motion magnitude.\label{fig:sub3}]{{\includegraphics[width=0.31\textheight,trim=30 2 30 50,clip,keepaspectratio]{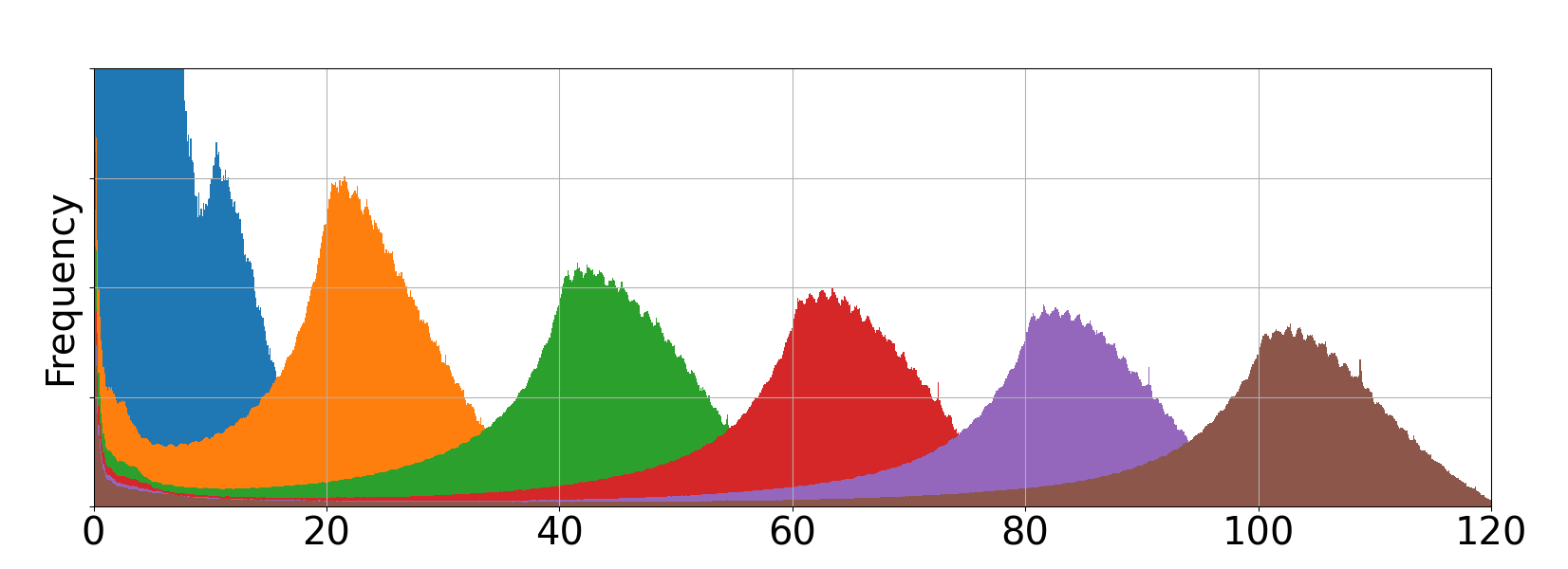} }}%
    \caption{Motion magnitude histograms of datasets Vimeo-90K (\ref{fig:sub1}), Xiph-4K (\ref{fig:sub2}) and Bracketed (\ref{fig:sub3}).}
    \label{fig:histograms}
    \vspace{-6ex}
\end{figure*}

We procedurally mine $512\mathord\times512$ image triplets from publicly available videos, extending the method described in~\cite{moblur-2019}. 
We apply this procedure to generate several motion brackets, %
i.e.: 0-20, 20-40, ..., 100-120. The motion distribution histograms of these brackets are shown overlapped in Figure~\ref{fig:histograms}. %
The effect of training using blends with increasing motion range is analyzed in Section~\ref{sec:study_motion_ranges}.

%% file: sec/4_implementation.tex
\section{Implementation Details}
\label{sec:implementation}
We implemented our model in TensorFlow 2. As training data, we use either Vimeo-90K or one of our large motion datasets described in Section~\ref{sec:training_datasets}. 

For the Vimeo-90K dataset, we use a batch size of 8, with a $256\mathord\times256$ random crop size, distributed over 8 NVIDIA V100 GPUs. We apply data augmentation: Random rotation with [-45$^{\circ}$,45$^{\circ}$], rotation by multiples of 90$^{\circ}$, horizontal flip, and reversing triplets. We use Adam~\cite{adam-optimizer} optimizer with $\beta_{1}\mathord=0.9$ and $\beta_{2}\mathord=0.999$, without weight decay. We use an initial learning rate of $1e^{-4}$ scheduled (piece-wise linear) with exponential decay rate of $0.464$, and decay steps of $750\mathrm{K}$, for $3\mathrm{M}$ iterations.

For comparison with the recent state-of-the-art models, we trained two versions: One optimized using $\mathcal{L}_{1}$ loss alone, which achieves higher benchmark scores, and another, that favours image quality, trained with our proposed style loss, $\mathcal{L}_{S}$. Our style loss optimally combines $\mathcal{L}_{1}$, $\mathcal{L}_{\rm VGG}$, and $\mathcal{L}_{\rm Gram}$. %

To perform our qualitative evaluations, we also implement the SoftSplat~\cite{softsplat-2020} in TensorFlow 2, since pre-trained models were not available at the time of writing. In the Supplementary Materials, we show our faithful implementations on a DAVIS~\cite{DAVIS} image sample rendered in~\cite{softsplat-2020}. We found that renderings with our implementation to be quite comparable to the ones provided in the original paper. We provide additional implementation details in the Supplementary.

%% file: sec/5_results.tex
\section{Results}
\label{sec:results}
Using existing benchmarks, we quantitatively compare FILM to recent methods: DAIN~\cite{dain-2019}, AdaCoF~\cite{adacof-2020}, BMBC~\cite{bmbc-2021}, SoftSplat~\cite{softsplat-2020}, and ABME~\cite{abme-2021}.  We additionally provide qualitative comparisons (to SoftSplat and ABME) on near-duplicate interpolation, for which no benchmarks currently exist.

\vspace{0.05in}
\noindent\textbf{Metrics.} We use common quantitative metrics: Peak Signal-To-Noise Ratio (PSNR) and Structural Similarity Image Metric (SSIM). High PSNR and SSIM scores indicate better quality.

\vspace{0.05in}
\noindent\textbf{Datasets.} 
We report metrics on Vimeo-90K~\cite{vimeo}, UCF101\cite{ucf101}, Middlebury~\cite{middlebury}, and on a 4K large motion dataset Xiph~\cite{xiph,softsplat-2020}. 
Figure~\ref{fig:histograms} shows motion magnitude histograms for Vimeo-90K and Xiph. Vimeo-90K (\ref{fig:sub1}) motion is limited to 25 pixels, while the Xiph (\ref{fig:sub2}) has a long-tailed distribution extending to 80 pixels.

In this comparison, all methods are with the Vimeo-90K dataset. To evaluate visual quality, we use a new challenging near-duplicate photos as the testing dataset. For ablation studies on large motion, we use our ``bracketed'' dataset (see Section~\ref{sec:training_datasets}) as the training dataset.

\subsection{Quantitative Comparisons}
\label{sec:quantitative_comparisons}
\vspace{0.05in}
\noindent\textbf{Small-to-Medium Motion.} \Table{medium_motion} shows midpoint frame interpolation comparisons with DAIN~\cite{dain-2019}, AdaCoF~\cite{adacof-2020}, BMBC~\cite{bmbc-2021}, SoftSplat~\cite{softsplat-2020}, and ABME~\cite{abme-2021} on small-to-medium motion datasets: Vimeo-90K, Middlebury, and UCF101. %

The SoftSplat method reports two sets of results, one set trained with color loss ($\mathcal{L}_{Lap}$), which performs better on standard benchmarks, and another trained with a perceptually-sensitive loss ($\mathcal{L}_{F}$), which leads to perceptually higher quality frames. The rest report results obtained by training with various color or low-level loss functions.
\input{tab/medium_motion}

Based on color losses, ABME outperforms all other methods on Vimeo-90K. On Middlebury and UCF101, SoftSplat trained with color loss has the highest PSNR. We note that ABME and SoftSplat are complex to train, each consisting of multiple sub-networks dedicated to motion estimation, refinement, or synthesis. Their training processes involve multiple datasets and stage-wise pre-training. Data scarcity, which is even more critical in large motion, could also complicate pre-training. Our unified, single-stage, FILM network achieves competitive PSNR scores.

The perception-distortion tradeoff~\cite{perception_tradeoff} proved that minimizing distortion metrics alone, like PSNR or SSIM, can have a negative effect on the perceptual quality. %
As such, we also optimize our model with our proposed Gram Matrix-based loss, $\mathcal{L}_{S}$, which optimally favours both color differences and perceptual quality. 

When including perceptually-sensitive losses, FILM outperforms the state-of-the-art SoftSplat on Vimeo-90K. We also achieve the highest scores on Middlebury and UCF101. In the next Subsection~\ref{subsec:qualitative_comparisons}, we show visual comparisons that support the quantitative gains in PSNR with gains in image quality.

\vspace{0.05in}
\noindent\textbf{Large Motion.} 
\Table{large_motion} presents midpoint frame interpolation comparisons on Xiph-2K and Xiph-4K, all methods (including FILM) trained on Vimeo-90K.  FILM outperforms all other models for color-based losses. Note that (not shown in the table) when training FILM on a custom large motion dataset, detailed in Section~\ref{sec:training_datasets}, we can achieve an additional performance gain of $+0.5$dB on Xiph-4K, the benchmark with the largest motions.

When including perceptually-sensitive losses, FILM outperforms SoftSplat-$\mathcal{L}_{F}$ in PSNR on Xiph-2K and both PSNR and SSIM on the larger motion Xiph-4K. Thus, FILM is better able to generalize from the small motions in the Vimeo-90K training datasets to the larger motions present in the Xiph test sets. We hope these findings will interest the greater research community, where large motion is often challenging. 
In the next Subsection~\ref{subsec:qualitative_comparisons}, we provide visual results that support the effectiveness of our method in samples with motion ranges as large as 100 pixels. 
\input{tab/large_motion}

\subsection{Qualitative Comparisons}
\label{subsec:qualitative_comparisons}
We provide visual results that support our quantitative results. We use the version of the model that yields high image quality, i.e.: our FILM-$\mathcal{L}_{S}$ and SoftSplat-$\mathcal{L}_{F}$. For ABME~\footnote{\url{https://github.com/JunHeum/ABME}}, we create visual results using the released pre-trained models. For SoftSplat\footnote{\url{https://github.com/sniklaus/softmax-splatting}\label{softmax}}~\cite{softsplat-2020}, we use our faithful implementation, since neither source code nor pre-trained model was publicly available at the time of writing. %

\begin{figure}[t!]
    \centering
    \subfloat[\centering Sharpness. \label{fig:sharpness}]{{\includegraphics[trim={16 0 16 0},clip,width=0.489\linewidth]{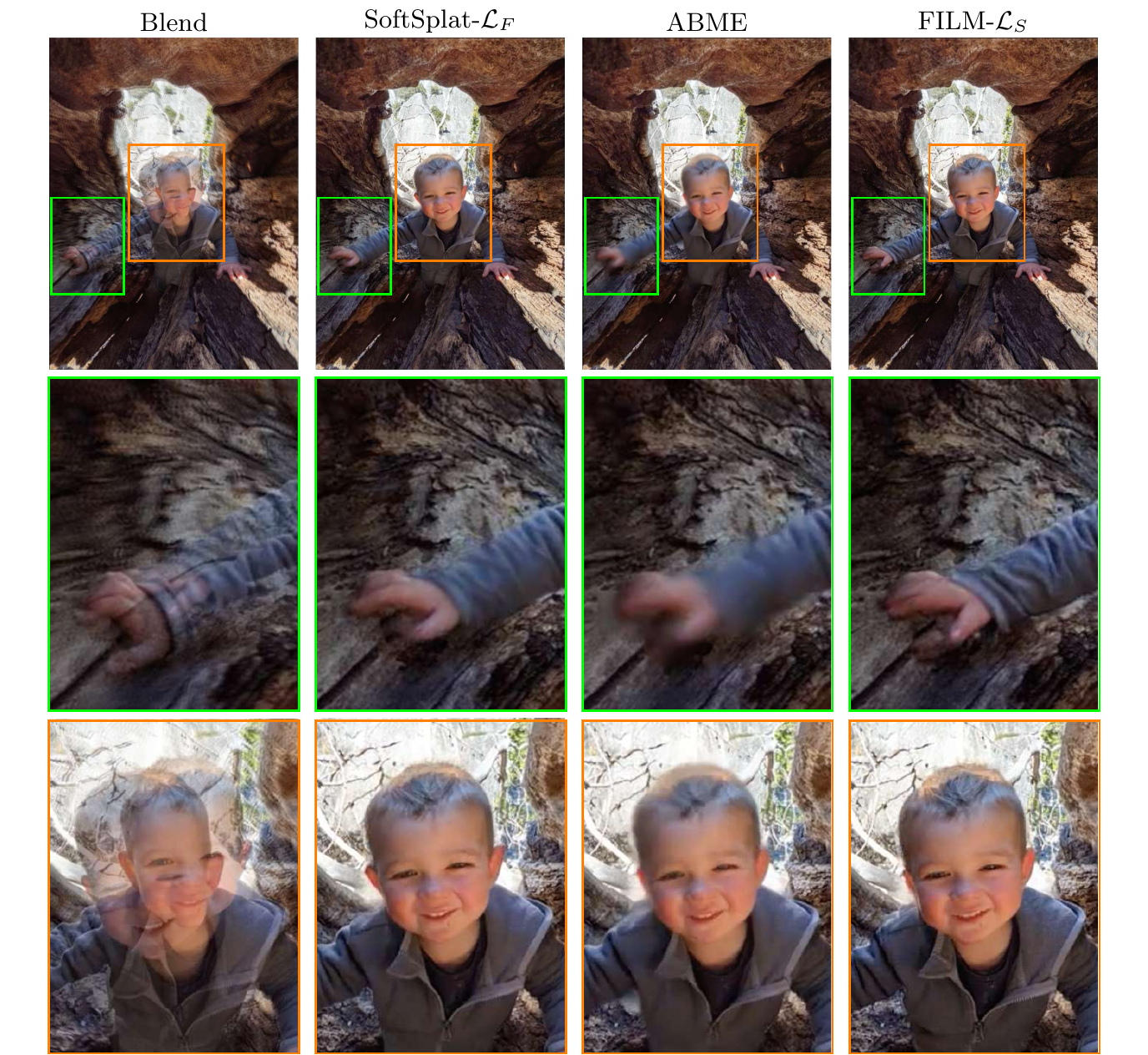} }}%
    \enspace
    \subfloat[\centering Inpainting disocclusions.\label{fig:disocclusions}]{{\includegraphics[trim={16 0 16 0},clip,width=0.489\linewidth]{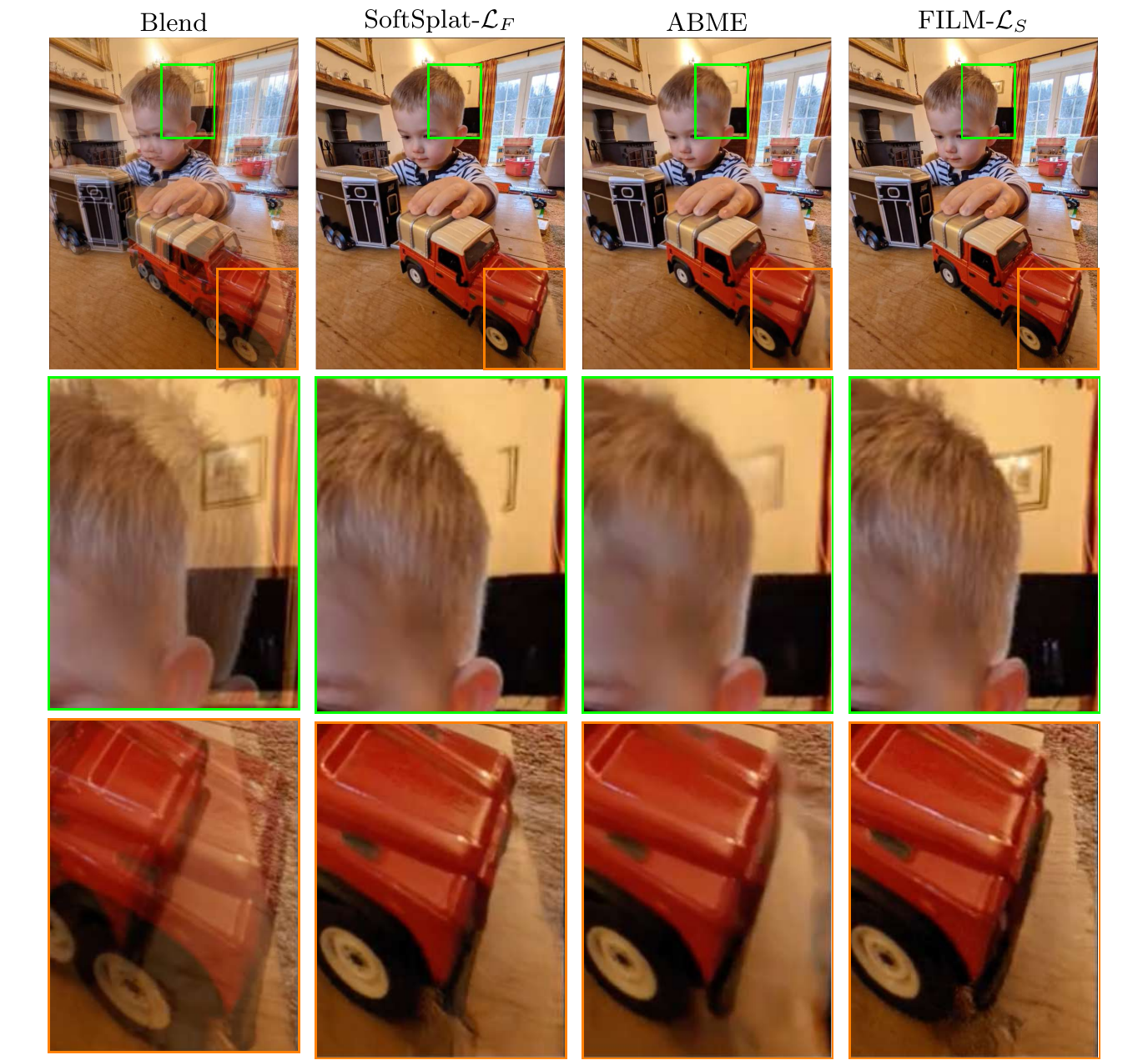} }}%
    \caption{Qualitative comparison on sharpness and large disocclusion inpainting. (\ref{fig:sharpness}) FILM produces sharp images and preserves the fingers. SoftSplat~\cite{softsplat-2020} shows artifacts (fingers) and ABME~\cite{abme-2021} has blurriness (the face). (\ref{fig:disocclusions}) Our method inpaints large disocclusions well, because of our proposed Gram Matrix-based (Style) loss. SoftSplat~\cite{softsplat-2020} and ABME~\cite{abme-2021} produce blurry inpaintings or un-natural deformations. }%
    \label{fig:qualitative}%
\end{figure}

\vspace{0.05in}
\noindent\textbf{Sharpness.} To evaluate the effectiveness of our Gram Matrix-based loss function (Equation~\ref{eq:gram-matrix}) in preserving image sharpness, we visually compare our results against images rendered with other methods.
As seen in Figure~\ref{fig:sharpness}, our method synthesizes visually superior results, with crisp image details on the face and preserving the articulating fingers.

\vspace{0.05in}
\noindent\textbf{Disocclusion Inpainting.} %
To effectively inpaint disoccluded pixels, models must learn appropriate motions or hallucinate novel pixels, this is especially critical in large scene motion, which causes wide disocclusions. Figure~\ref{fig:disocclusions} shows different methods, including ours, inpainting large disocclusions. Compared to the other approaches, FILM correctly paints the pixels while maintaining high fidelity. It also preserves the structure of objects, e.g. the red toy car, while SoftSplat~\cite{softsplat-2020} shows deformation, and ABME~\cite{abme-2021} creates blurry inpainting.

\begin{figure*}[t!]
    \centering
    \includegraphics[trim={16 0 16 0},clip,width=1.0\linewidth]{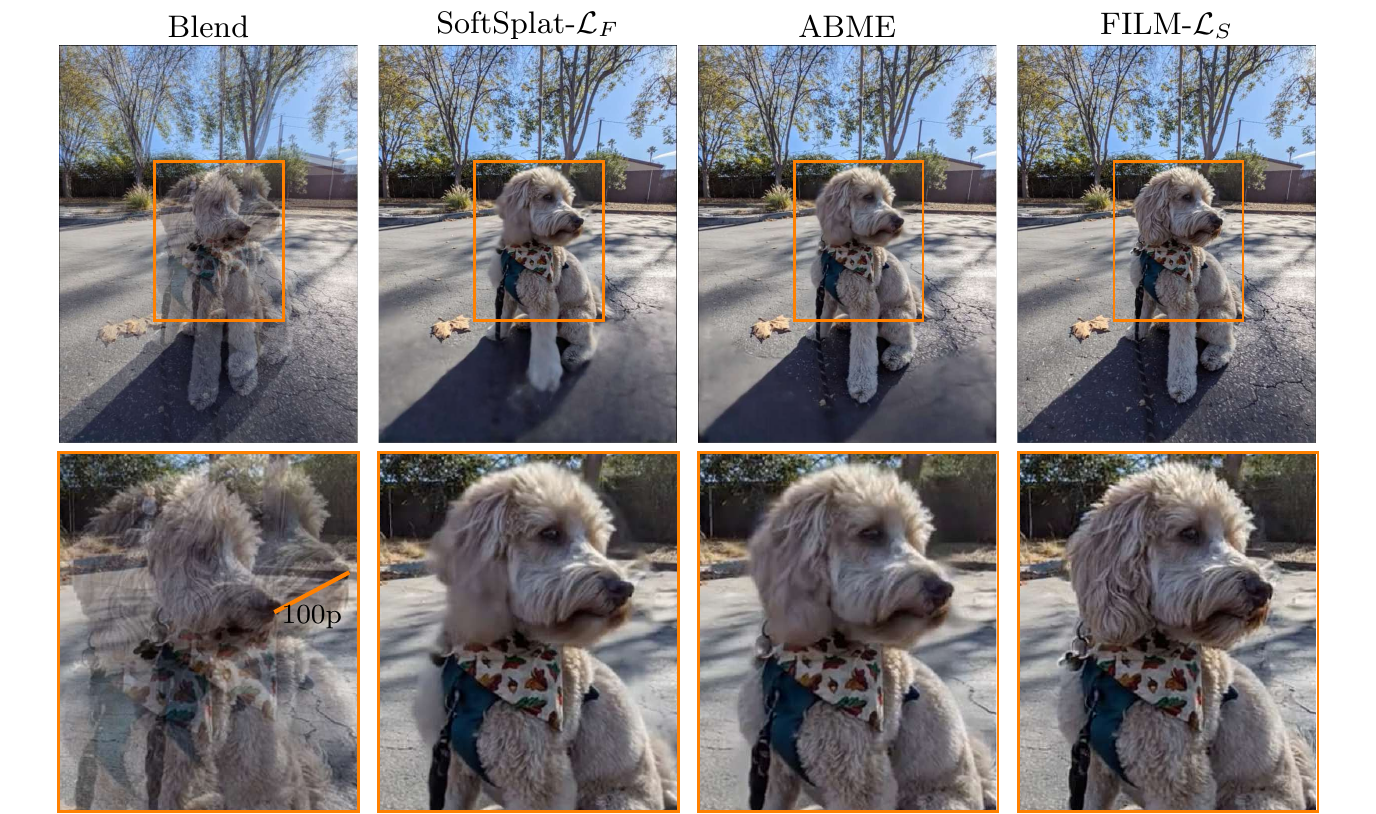}
    \caption{Qualitative comparison on large motion. Inputs with ~100pixels disparity overlaid (left). Although both SoftSplat~\cite{softsplat-2020}, ABME~\cite{abme-2021} capture the motion on the dog's nose, they appear blurry, and create a large artifact on the ground. FILM's strength is seen capturing the motion well and maintaining the background details.
    }
    \label{fig:large_motion}
    \vspace{-4ex}
\end{figure*}

\vspace{0.05in}
\noindent\textbf{Large Motion.} Large motion is one of the most challenging aspects of frame interpolation. %
Figure~\ref{fig:large_motion} shows results for different methods on a sample with 100 pixels disparity. Both SoftSplat~\cite{softsplat-2020} and ABME~\cite{abme-2021} were able to capture motions near the dog's nose, however they create large artifacts on the ground. FILM's strength is seen capturing the motion well and keeping the background details. Please see our Supplementary Materials for more visual results. %

\begin{figure*}
\centering
\setlength{\tabcolsep}{1.5pt}
\begin{tabular}{ccc}
\adjincludegraphics[height=4.2cm,trim={{.3\width} {.2\height} {.3\width} {.2\height}}, clip]{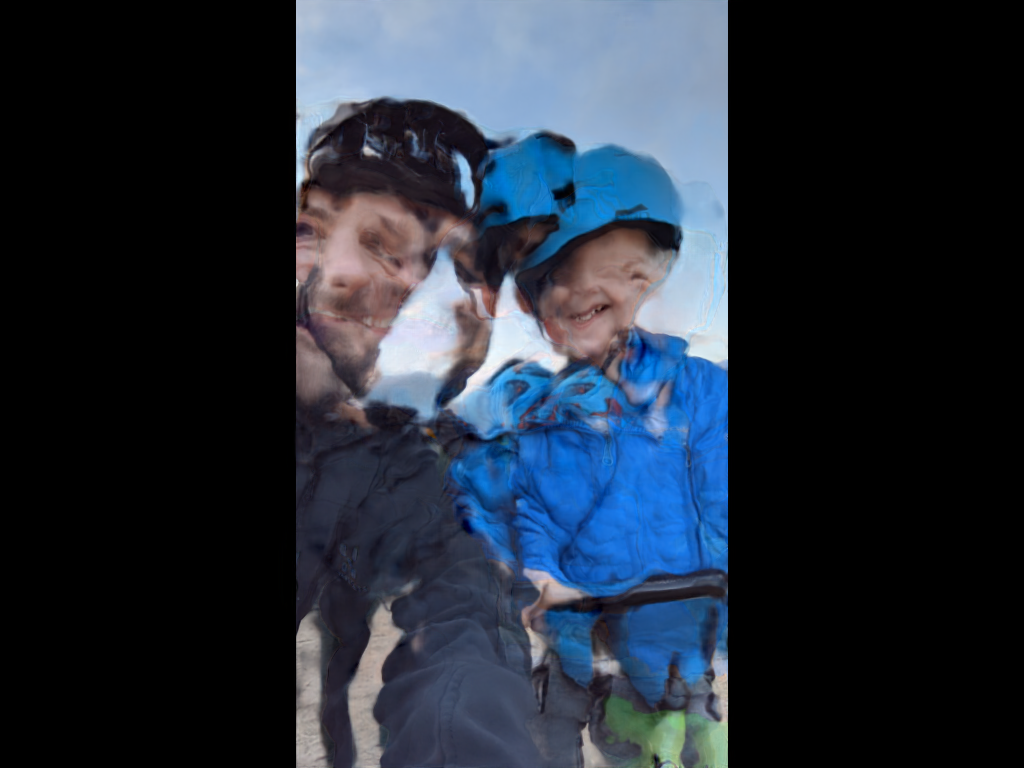} &
\adjincludegraphics[height=4.2cm,trim={{.3\width} {.2\height} {.3\width} {.2\height}}, clip]{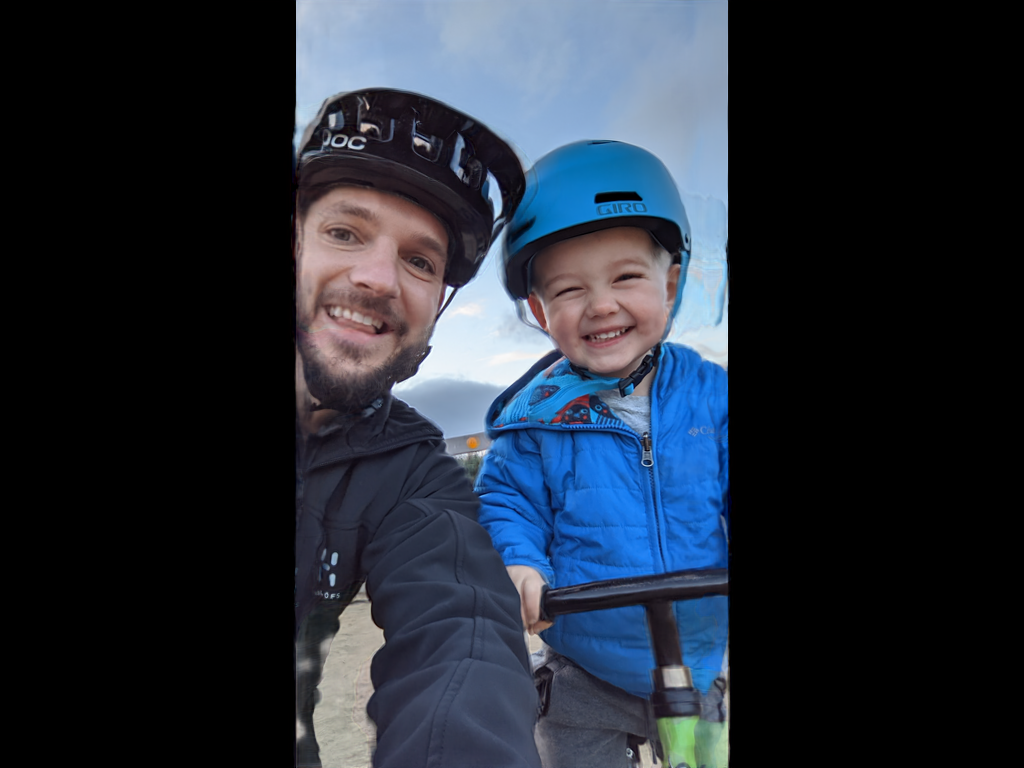} &
\adjincludegraphics[height=4.2cm,trim={{.3\width} {.2\height} {.3\width} {.2\height}}, clip]{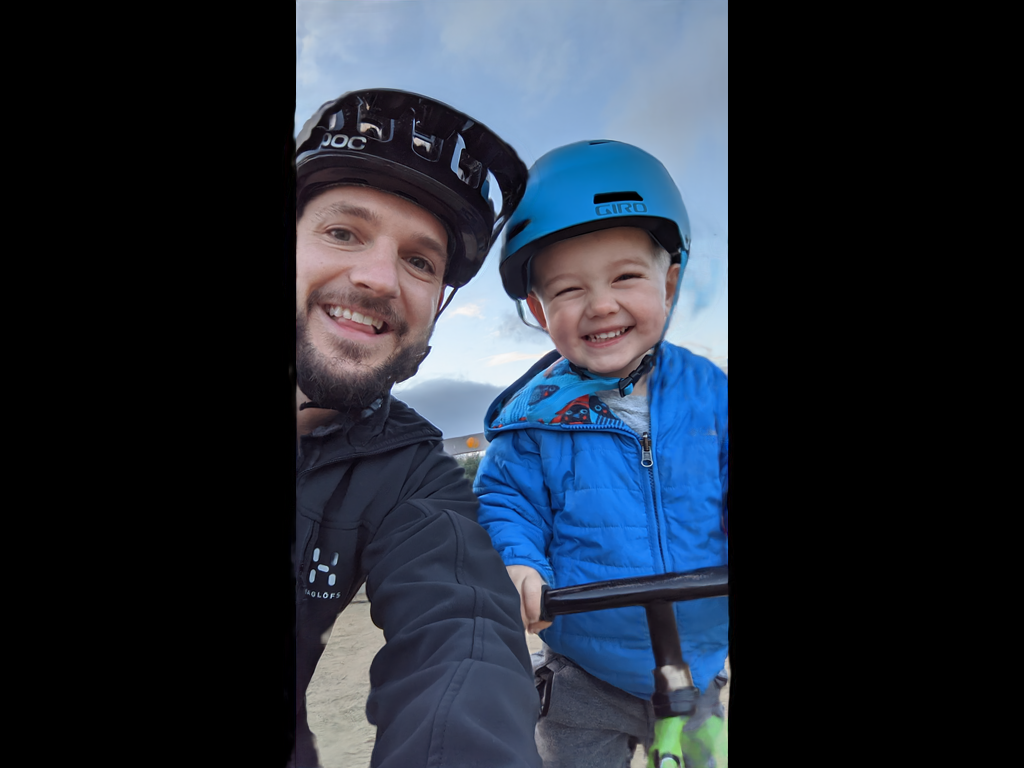} 
\end{tabular}
\caption{The visual impact of sharing weights. {\bf Left}: no sharing (FILM-med), {\bf Middle}: sharing (FILM-med), {\bf Right}: sharing (FILM), i.e., the highest quality  model.  Sharing weights is clearly better. Quality and sharpness increases when going from the medium to the full model with sharing.}
\vspace{-0.12in}
\label{fig:sharing_qualitative}
\end{figure*}

\subsection{Ablations}
\label{sec:ablation_sharing}
In this section, we present ablation studies to analyze the design choices of FILM.

\vspace{0.05in}
\noindent\textbf{Weight Sharing.} We compare our shared feature extractor with a regular UNet-like encoder~\cite{unet} that uses independent weights at all scales, forcing us to also learn independent flows. Table~\ref{tab:my_label} presents mid-frame results in PSNR.
It is not straightforward to construct models that are fair to compare: one could either match the total number of weights or the number of filters at each level. 
We chose to use a UNet encoder that starts from the same number of filters as ours and then doubles the number at each level. The FILM-model we have used in this paper starts with 64 filters, so this leads to a UNet with feature counts: $[64,128,256...]$. We find that training with this configuration does not converge without weight sharing. To study this further, we construct two simpler variants of our model, starting from 32 filters. We are able to train these two models with a small loss in PSNR as compared to the equivalent model that shares weights. 

To conclude, weight sharing allows training a more powerful model, reaching a higher PSNR. Additionally, sharing may be important for fitting models in GPU memory in practical applications. Further, the model with weight sharing is visually superior with substantially better generalization when testing on pairs with motion magnitude beyond the range in the training data (see Figure~\ref{fig:sharing_qualitative}).

\begin{table}
    \centering
    \begin{tabular}{l|c|c|}
    \bf{model} & \bf{PSNR (w/ sharing)} & \bf {PSNR (w/o sharing)} \\
    \hline
    FILM                &  36.06 & N/A \\
    FILM-med            &  35.30 & 35.28 \\
    FILM-lite           &  35.09 & 34.93 \\
    \end{tabular}
    \caption{Weight sharing ablation study. A model without multi-scale feature sharing achieves results that are slightly lower than those achieved with shared features, e.g. FILM-med and FILM-lite. We have not been able to train our highest quality model (FILM) without weight sharing as the training gets unstable (indicated with N/A).}
    \label{tab:my_label}
    \vspace{-6ex}
\end{table}
    
\begin{figure*}[ht!]
    \centering
    \includegraphics[trim={9 139 0 0},clip,width=1.0\linewidth]{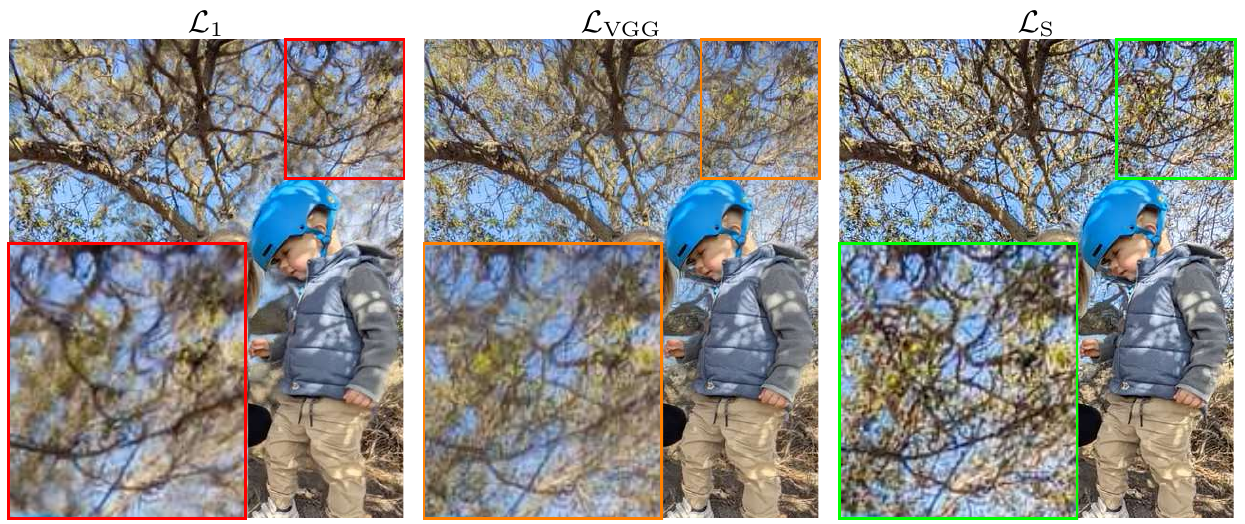}
    \caption{Loss function comparison on our FILM. L1 loss (left), L1 plus VGG loss (middle), and our proposed style loss (right), showing significant sharpness improvements (green box).
    }
    \label{fig:loss_ablation}
    \vspace{-2ex}
\end{figure*}

\noindent\textbf{Gram Matrix (Style) Loss.} Figure~\ref{fig:loss_ablation} presents qualitative results of FILM trained with $\mathcal{L}_{1}$, adding $\mathcal{L}_{VGG}$, and with our proposed style loss $\mathcal{L}_{S}$, given by Equation~\ref{eq:style}. Using $\mathcal{L}_{1}$ alone leads to blurry images (red box), while adding $\mathcal{L}_{VGG}$ loss reduces blurry artifacts (orange box). Our proposed loss (green box) significantly improves sharpness of our FILM method.

\begin{wrapfigure}{r}{0.5\textwidth}
 \vspace{-10ex}
  \begin{center}
    \includegraphics[width=0.5\textwidth,trim=1 0 0 10,clip]{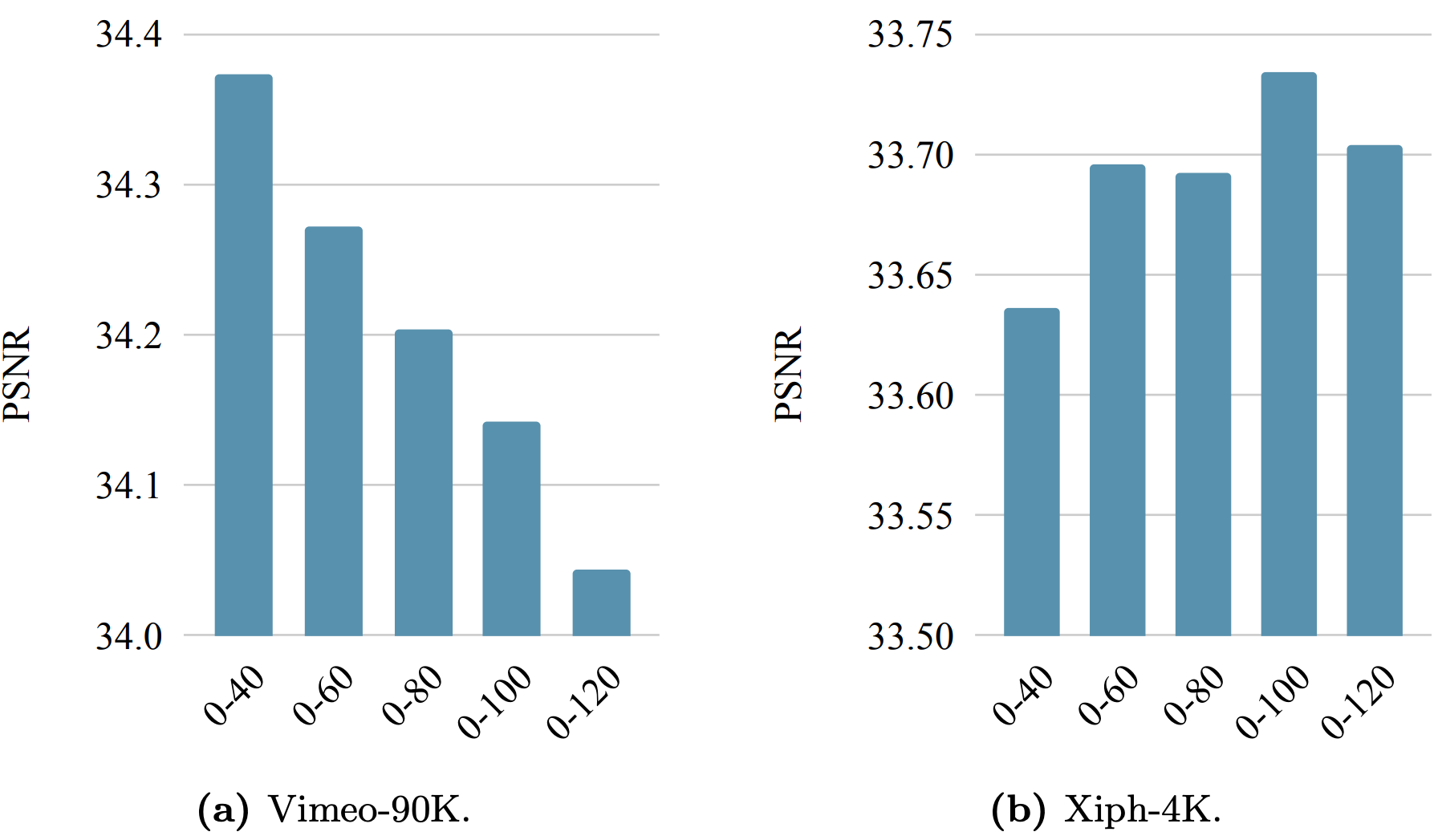}
  \end{center}
  \caption{A study of the effect of the training dataset's motion range on the PSNR, when evaluated on (a)~Vimeo-90K, and (b)~Xiph-4K. %
  }
  \label{fig:motion_ranges}
  \vspace{-10ex}
\end{wrapfigure}

\vspace{0.05in}
\noindent\textbf{Motion Ranges.} We study the effect of the training dataset motion range on the model's ability to handle different motions at test time, using the ``bracketed dataset'' (see Section~\ref{sec:training_datasets}). For this study, we use a reduced FILM-med to save compute resources. %
FILM-med is trained on motion ranges of 0-40, 0-60, 0-80, 0-100 and 0-120 pixels. They are evaluated on Vimeo-90K (0-25 range) and Xiph\-4K (0-80 range), as shown in Figure~\ref{fig:motion_ranges}.
On Vimeo-90K, FILM-med trained with the smallest motion range performs the best. As larger motion is added, PSNR goes down significantly. 

We hypothesize that this behavior is caused by two factors: 1) when training for larger motion, the model assumes a larger search window for correspondence, and thus has more chances to make errors. 2) with larger motion, the problem is simply harder and less neural capacity is left for smaller motion.

On Xiph-4K, the best performance is obtained by training with motion range 0-100 pixels. Motivated by our finding, we trained our best model (FILM) with this dataset, and achieve an additional PSNR performance gain of +0.5dB over the state-of-the-art, as described in Section~\ref{sec:quantitative_comparisons}.

In summary, our findings indicate that scale-agnostic features and shared weight flow prediction improve the model's ability to learn and generalize to a wider range of motion. In addition, we find that best results are obtained when the training data also matches the test-time motion distribution.
\label{sec:study_motion_ranges}

\subsection{Performance and Memory}
Table~\ref{tab:computational_complexity} presents inference times and memory comparisons on an NVIDIA V100 GPU. We report the average of 100 runs. Our FILM is \textbf{$3.95\times$} faster than ABME, and \textbf{$9.75\times$} faster than SoftSplat, while using only \textbf{$1.27\times$} and \textbf{$1.01\times$} more memory than ABME and SoftSplat, respectively. Our model is slightly larger due to its deep pyramid (7 levels), but the time performance gains are significantly large.

\subsection{Limitations}
In some instances, FILM produces un-natural deformations when the in-between motion is extreme. Although resulting videos are appealing, the subtle movements may not look natural. We provide failure examples in our Supplementary video.

\begin{table}[h!]
\small
\centering
\scalebox{0.85}{
\begin{tabular}{l c c}
    \toprule
    \shortstack{Interpolation\\Method} & \shortstack{Inference Time \\ (Second)$\downarrow$} &
    \shortstack{Peak Memory \\ (GB)$\downarrow$} \\
    \hline
     SoftSplat & 3.834 & 4.405 \\
     ABME & 1.554 & \bf{3.506} \\
     Our FILM & \bf{0.393} & 4.484 \\
    \bottomrule
\end{tabular}
}
\caption{Inference time and memory comparison for a $720$p frame interpolation.}
\label{tab:computational_complexity}
\vspace{-8ex}
\end{table}

%% file: tab/medium_motion.tex
\begin{table}[t!]
\small
\renewcommand\tabcolsep{3.0pt}
\centering
\resizebox{\linewidth}{!}{ %
\begin{tabular}{lcc@{\hskip 0.5cm}cc@{\hskip 0.5cm}cc}
\toprule
&  \multicolumn{2}{c}{\textbf{Vimeo-90K}~\cite{vimeo}}{\hskip 0.5cm} & \multicolumn{2}{c}{\textbf{Middlebury}~\cite{middlebury}}{\hskip 0.5cm}  & \multicolumn{2}{c}{\textbf{UCF101}~\cite{ucf101}}{\hskip 0.25cm} \\
\cline{2-7}
 & PSNR$\uparrow$ & SSIM$\uparrow$ & PSNR$\uparrow$ & SSIM$\uparrow$ & PSNR$\uparrow$ & SSIM$\uparrow$ \\ 
\hline
DAIN & 34.70 & 0.964 & 36.70 & 0.965 & 35.00 & 0.950 \\ 
AdaCoF & 34.35 & 0.973 & 35.72 & \textbf{\textcolor{blue}{0.978}} & 34.90 & 0.968 \\ 
BMBC & 35.01 & 0.976 & n/a & n/a & 35.15 & 0.969 \\ 
SoftSplat-$\mathcal{L}_{Lap}$ & 36.10 & 0.970 & \textbf{\textcolor{blue}{38.42}} & 0.971 & \textbf{\textcolor{blue}{35.39}} & 0.952 \\ 
ABME & \textbf{\textcolor{blue}{36.18}} & \textbf{\textcolor{blue}{0.981}} & n/a & n/a & 35.38 & \textbf{\textcolor{blue}{0.970}} \\ 
Our FILM-$\mathcal{L}_{1}$ & 36.06 & 0.970 & 37.52 & 0.966 & 35.32 & 0.952 \\ 
\hline
SoftSplat-$\mathcal{L}_{F}$ & 35.48 & 0.964 & 37.55 & 0.965 & 35.10 & 0.948 \\
Our FILM-$\mathcal{L}_{\rm VGG}$ & 35.76 & 0.967 & 37.43 & \textbf{\textcolor{red}{0.966}} & \textbf{\textcolor{red}{35.20}} & \textbf{\textcolor{red}{0.950}} \\ 
Our FILM-$\mathcal{L}_{S}$ & \textbf{\textcolor{red}{35.87}} & \textbf{\textcolor{red}{0.968}} & \textbf{\textcolor{red}{37.57}} & \textbf{\textcolor{red}{0.966}} & 35.16 & 0.949 \\ 
\bottomrule
\end{tabular}
} %
\caption{Comparison on {\em small-to-medium motion} benchmarks. Best scores for color losses are in \textbf{\textcolor{blue}{blue}},  for perceptually-sensitive losses in \textbf{\textcolor{red}{red}}. In this comparison, all methods are trained on Vimeo-90K.}
\label{tab:medium_motion}
\vspace{-4ex}
\end{table}

%% file: tab/large_motion.tex
\begin{table}[t!]
\small
\renewcommand\tabcolsep{3.0pt}
\centering
\begin{tabular}{lcc@{\hskip 0.5cm}cc}
\toprule
&  \multicolumn{2}{c}{\textbf{Xiph-2K}~\cite{xiph}}{\hskip 0.5cm} & \multicolumn{2}{c}{\textbf{Xiph-4K}~\cite{xiph}}{\hskip 0.5cm} \\
\cline{2-5}
 & PSNR$\uparrow$ & SSIM$\uparrow$ & PSNR$\uparrow$ & SSIM$\uparrow$ \\ 
\hline
DAIN & 35.95 & 0.940 & 33.49 & 0.895 \\ 
ToFlow & 33.93 & 0.922 & 30.74 & 0.856 \\
AdaCoF & 34.86 & 0.928 & 31.68 & 0.870 \\ 
BMBC & 32.82 & 0.928 & 31.19 & 0.880 \\ 
ABME & 36.53 & 0.944 & 33.73 & 0.901 \\ 
SoftSplat-$\mathcal{L}_{Lap}$ & 36.62 & 0.944 & 33.60 & 0.901 \\ 
Our FILM-$\mathcal{L}_{1}$ & \textbf{\textcolor{blue}{36.66}} & \textbf{\textcolor{blue}{0.951}} & \textbf{\textcolor{blue}{33.78}} & \textbf{\textcolor{blue}{0.906}} \\ 
\hline
SoftSplat-$\mathcal{L}_{F}$ & 35.74 & \textbf{\textcolor{red}{0.944}} & 32.55 & 0.865 \\
Our FILM-$\mathcal{L}_{S}$ & \textbf{\textcolor{red}{36.38}} & 0.942 & \textbf{\textcolor{red}{33.29}} & \textbf{\textcolor{red}{0.882}} \\ 
\bottomrule
\end{tabular}
\caption{Comparison on {\em large motion} benchmarks. Best scores for color losses in (\textbf{\textcolor{blue}{blue}}), and for perceptually-sensitive losses in (\textbf{\textcolor{red}{red}}). In this comparison, all methods are trained on Vimeo-90K}.
\label{tab:large_motion}
\vspace{-4ex}
\end{table}

%% file: sec/6_conclusions.tex
\section{Conclusions}

We have introduced an algorithm for large motion frame interpolation (FILM), in particular, for near-duplicate photos interpolation. FILM is a simple, unified and single-stage model, trainable from regular frames, and does not require additional optical-flow or depth prior networks, or their scarce pre-training data. Its core components are a feature pyramid that shares weight across scales and a ``scale-agnostic'' bi-directional motion estimator that learns from frames with normal motion but generalizes well to frames with large motion. To handle wide disocclusions caused by large motion, we optimize our models with the Gram matrix loss that matches the correlation of features to generate sharp frames. Extensive experimental results show that FILM outperforms other methods on large motions while still handling small motions well, and generates high quality and temporally smooth videos. Source codes and pre-trained models are available at \url{https://film-net.github.io}.

%% file: supplementary.tex
\appendix
\title{ {\sc Supplementary Materials }\\FILM: Frame Interpolation for Large Motion}  

\titlerunning{FILM: Frame Interpolation for Large Motion}
\authorrunning{F. Reda et al.} 
\titlerunning{FILM: Frame Interpolation for Large Motion}
\author{Fitsum Reda\inst{1} \and
Janne Kontkanen\inst{1} \and
Eric Tabellion\inst{1} \and
Deqing Sun\inst{1} \and \\
Caroline Pantofaru\inst{1} \and
Brian Curless\inst{1,2}}
\authorrunning{F. Reda et al.}
\institute{\hspace{-1ex}Google Research \and \hspace{-1ex}University of Washington}

\maketitle

We provide additional implementation details and a supplementary video (visit \url{https://film-net.github.io}) for a quick overview of our paper, the motivations, as well as more visual results. 

\section{Implementation Details}
\subsection{Loss Combination Weights}
Our style loss ($\mathcal{L}_{\rm{S}}$) optimally combines $\mathcal{L}_{1}$, $\mathcal{L}_{\rm VGG}$, and $\mathcal{L}_{\rm Gram}$. We use a piecewise linear weight-schedule to select a weight for each loss at each iteration. Specifically, we assign weights of $(1.0, 1.0, 0.0)$ for $1.5\mathrm{M}$ iterations, and weights of $(1.0, 0.25, 40.0)$ for the last $1.5\mathrm{M}$ iterations. For the last $1.5\mathrm{M}$ iterations, the loss weights are empirically selected such that each loss contributes equally to the combined Style loss.

\subsection{SoftSplat Implementation}
As described in the paper, we have implemented the SoftSplat[18] in Tensorflow 2, using the author's tuned hyper-parameters, and we have been able to replicate their published benchmark scores. Note that, SoftSplat\footnote{\url{https://github.com/sniklaus/softmax-splatting}} provides only a CuPy implementation of the softmax splatting operator.

\begin{figure}[t!]
\centering
    \includegraphics[trim={30 0 30 0},clip,width=1.0\linewidth]{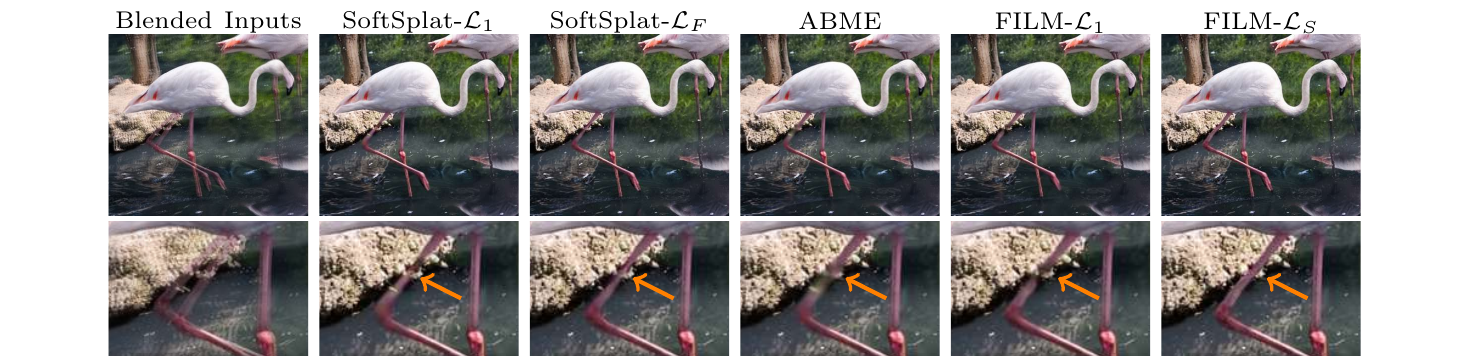}
    \vspace{-4ex}
    \caption{
    Frame interpolation example from DAVIS-dataset~[24] featuring a walking flamingo. From left to right: The input frames overlaid, SoftSplat-${\mathcal{L}_{1}}$~[18], SoftSplat-${\mathcal{L}_{F}}$~[18], ABME~[23], our FILM-${\mathcal{L}_{1}}$, and our FILM-${\mathcal{L}_{S}}$. FILM-${\mathcal{L}_{S}}$ produces crisp frame, while color distortions and transparencies can be seen in ABME and SoftSplat, respectively. SoftSplat renderings are generated from our faithful implementations, which we found to be quite comparable to the original paper~[18].}
    \label{fig:flamingo}
\end{figure}

In Figure~\ref{fig:flamingo}, we show our faithful implementations on a DAVIS~[34] image sample rendered in~[18]. We found that renderings with our implementation to be quite comparable to the ones provided in the original paper. 

\section{Supplementary Video}
Please watch the enclosed video or visit \url{https://film-net.github.io}. We have included motivations, illustrations of our methods, and more visual results and failure samples.